\documentclass{article}

\usepackage{microtype}
\usepackage{graphicx}
\usepackage{subfigure}
\usepackage{booktabs} 

\usepackage[utf8]{inputenc} 
\usepackage[T1]{fontenc}    

\usepackage{hyperref}

\usepackage[accepted]{icml2024}

\usepackage{amsmath}
\usepackage{amssymb}
\usepackage{mathtools}
\usepackage{amsthm}
\usepackage{graphicx}
\usepackage{subcaption}
\usepackage{caption}
\usepackage[capitalize,noabbrev]{cleveref}

\usepackage{url}            
\usepackage{amsfonts}       
\usepackage{nicefrac}       
\usepackage{microtype}      
\usepackage{xcolor}         

\theoremstyle{plain}

\theoremstyle{definition}

\theoremstyle{remark}

\usepackage[textsize=tiny]{todonotes}

\usepackage[labelformat=simple]{subcaption}

\usepackage{placeins}

\usepackage{graphbox}
\usepackage{soul}
\usepackage{tcolorbox}
\usepackage{paralist}
\usepackage{multirow}

\DeclareRobustCommand{\E}{\operatornamewithlimits{\mathbb{E}}}

\newdimen\indexdigits
\setbox0\hbox{9999}
\indexdigits\wd0
\newcommand\paddigits[1]{\hbox to \indexdigits{\hfill#1}}

\icmltitlerunning{Diffusion Domain Expansion: Learning to Coordinate Pre-trained Diffusion Models}

\begin{document}

\twocolumn[
\icmltitle{Diffusion Domain Expansion: Learning to Coordinate\\Pre-Trained Diffusion Models}



\icmlsetsymbol{equal}{*}

\begin{icmlauthorlist}
\icmlauthor{Egor Lifar}{equal,yyy}
\icmlauthor{Semyon Savkin}{equal,yyy}
\icmlauthor{Timur Garipov}{yyy}
\icmlauthor{Shangyuan Tong}{yyy}
\icmlauthor{Tommi Jaakkola}{yyy}
\end{icmlauthorlist}

\icmlaffiliation{yyy}{MIT CSAIL, Cambridge, USA}

\icmlcorrespondingauthor{Egor Lifar}{egor.lifar@gmail.com}

\icmlkeywords{Machine Learning, ICML}

\vskip 0.3in
]



\printAffiliationsAndNotice{\icmlEqualContribution} 

\begin{abstract}
In this paper, we propose Diffusion Domain Expansion (DDE), a method that efficiently extends pre-trained diffusion models to generate larger objects and handle more complex conditioning beyond their original capabilities. Our method employs a compact trainable network designed to coordinate the denoised outputs of pre-trained diffusion models. We demonstrate that the coordinator can be universally simple while being capable of generalizing to domains larger than those observed during its training time. We evaluate DDE on long audio track generation and conditional image generation, demonstrating its applicability across domains. DDE outperforms other approaches to coordinated generation with diffusion models in qualitative and quantitative evaluations.
\end{abstract}

\section{Introduction}
\label{sec:introduction}

Diffusion models have delivered state-of-the-art results in generating images \cite{betker2023improving}, music \cite{evans2024fast}, robot control trajectories \cite{chi2023diffusionpolicy}, and molecular structures \cite{abramson2024accurate}. As training at scale incurs significant costs, research on the control and re-use of pre-trained models has gained substantial attention. Post-training tasks include conditional generation \cite{sohl2015deep, zhang2023adding}, in-painting \cite{choi2021ilvr}, editing \cite{meng2022sdedit}, and composition of models~\cite{du2023reduce, garipov2023compositional}\footnote{See Appendix \ref{sec:related} for a detailed review of related work.}.

A recent line of work \cite{bar2023multidiffusion, lee2023syncdiffusion, zhang2023diffcollage, wu2024compositional} focuses on expanding the generative domain of diffusion models to generate data samples with multiple parts (e.g., panoramic images) by coordinating pre-trained diffusion processes. These approaches prescribe fixed algorithmic coordination mechanisms, which
do not require any additional training and can be applied off-the-shelf. They are, however, limited to specific tasks they are designed for. \citet{huang2023collaborative} proposed to train a dynamic diffuser module to fuse unimodal diffusion models for multimodal conditioning. This method expands the generative domain by increasing the number of conditioning inputs but requires the modalities to be fixed at training time.

In this paper, we propose Diffusion Domain Expansion (DDE), a method for efficiently training small coordinators that extend the generative domain of pre-trained models along two axes: the size of the generated object (e.g., larger images) and the size of the conditioning object (e.g., multiple labels). We demonstrate the ability of coordinators to generalize to domains larger than those seen during training.
We evaluate DDE on music and conditional image generation tasks, including CLEVR scene generation and satellite image generation. We show that our ViT-based coordinator delivers high-quality results by learning long-range dependencies while enforcing local consistency via MultiDiffusion-like updates.

\section{Background}
\label{sec:background}


Given a training data distribution $p_\text{data}(x)$ over $\mathbb{R}^d$, a diffusion process $p(x; t)$ is a time-indexed collection of distributions such that $p(x; t) = \int \mathcal{N}(x; x_0, \sigma^2(t)) p_\text{data}(x_0)\, dx_0$ is constructed by adding Gaussian noise with standard deviation $\sigma(t)$ to the ``clean'' samples from the data distribution ($x(t) = x_0 + \varepsilon; \varepsilon \sim \mathcal{N}(0, \sigma^2(t))$). In the following, we set $\sigma(t) = t$ and consider $t \in [t_\text{min}, t_\text{max}]$. We refer the reader to \cite{karras2022elucidating} for the discussion of other possible specifications of diffusion processes.

A diffusion model learns a denoising function $D: \mathbb{R}^d \times [t_\text{min}, t_\text{max}] \to \mathbb{R}^d$ which estimates the ``clean'' object given its noisy observation at time $t$. This function is trained by minimizing the denoising error
\begin{equation}%
    \label{eq:denoising}
    \E\limits_{t \sim p(t)} \E\limits_{x \sim p_\text{data}(x)} \E\limits_{\varepsilon \sim \mathcal{N}(0, t^2)}\Big[
        \lambda(t) \big\|D(x + \varepsilon, t) - x\big\|_2^2
    \Big],
\end{equation}%
where $p(t)$ is a distribution over $[t_\text{min}, t_\text{max}]$ and $\lambda(t) > 0$. 

The denoising function that minimizes \eqref{eq:denoising} is connected to the score $\nabla_x \log p(x, t)$ of the diffusion process:%
\begin{equation}
    \label{eq:score_via_D}
    \nabla_x \log p(x, t) = \big(D(x, t) - x\big) \,/ \,{t^2}.
\end{equation}%
It can be shown \cite{anderson1982reverse, song2021scorebased, karras2022elucidating} that, given access to the score $\nabla_x \log p(x, t)$, the distributions $p(x, t)$ can be recovered via the probability flow induced by the backward ODE%
\begin{equation}%
    \label{eq:backward_ode}
    dx = - t ~ \nabla_x \log p(x, t)\,dt,
\end{equation}%
which is initialized with $x(t_\text{max}) \sim p(x, t_\text{max})$ and runs backwards in time from $t=t_\text{max}$ to $t = t_\text{min}$. 

If $t_\text{max}$ is large enough  $p(x, t_\text{max})$ can be accurately approximated by a Gaussian prior $p_\text{prior}(x) = \mathcal{N}(0, t_\text{max}^2)$. After $D(\cdot, \cdot)$ is trained with the objective \eqref{eq:denoising}, the samples can be generated by sampling $x(t_\text{max}) \sim p_\text{prior}$ and numerically integrating the backward ODE \eqref{eq:backward_ode} with the score \eqref{eq:score_via_D}.


\section{Method}
\label{sec:method}

\subsection{General formulation}
\label{ssec:method_general}

Consider a pre-trained conditional diffusion model $p(x \vert y)$ with the denoising network $D(x, y, t)$. 
We are interested in expanding the generative domain, i.e., solving the conditional generation $p(X_{[L]} \vert Y_{[L]})$, where $X_{[L]}$ and $Y_{[L]}$ are the expanded generated object (e.g., a larger image) and the expanded conditioning input (e.g., a larger conditioning image). Naturally, we want to re-use the pre-trained model $p(x \vert y)$ and realize the generation in the expanded domain by employing the knowledge captured by $D(x, y, t)$. We utilize the assumption that the expanded output-input pair $(X_{[L]}, Y_{[L]})$ can be decomposed into a collection of smaller parts $([x_1, \ldots, x_L], [y_1, \ldots, y_L]) = F(X_{[L]}, Y_{[L]})$, where all $x_i \in \mathcal{X}$ and $y_i \in \mathcal{Y}$ are in the respective base domains and $L$ denotes the number of smaller parts. The mapping $F(\cdot, \cdot)$ defines the decomposition process. For instance, in image generation $X_{[L]}$ corresponds to a larger image from which we can extract a set of $L$ smaller same-sized patches $[x_1, \ldots, x_L]$ that cover the large image completely. In our framework, the patches are allowed to have overlaps.

Each of the smaller objects $x_i$ can be generated using $p(x_i \vert y_i)$. However, to generate a coherent large object $X_{[L]}$, the generative processes need to be coordinated. The idea of our approach is to train a coordinator network $C_{[\cdot]}$ that learns to coordinate diffusion processes by operating on the outputs of the pre-trained denoising networks $[D(x_i, y_i, t)]_{i=1}^L$. We use a parameter-efficient architecture for the coordinator network $C_{[\cdot]}$ and we train it on a dataset of larger objects $\mathcal{D}_{L_\text{train}}^N = \{(X^{(i)}_{[L_\text{train}]}, Y^{(i)}_{[L_\text{train}]})\}_{i=1}^N$ of size $L_\text{train}$. Crucially, we demonstrate that after being trained on examples of size $L_\text{train}$, the coordinated diffusion $p(X_{[\cdot]} \vert Y_{[\cdot]})$ induced by $C_{[\cdot]}$ can generalize and generate objects of larger sizes $L_\text{test} \geq L_\text{train}$. We describe the general formulation and the training objective of DDE below, and we describe the coordinator architecture in Section~\ref{ssec:architecture}.

The DDE generative process is the reverse of the noising process $X_{[L]}(t) \sim \mathcal{N}(X_{[L]}(0), t^2)$ over expanded data $X_{[L]}$. For this expanded process, we build a composite denoiser $D_{[\cdot]}$ which utilizes the decomposition function $F$, the pre-trained base domain denoiser network $D$, and a trainable small-scale coordinator network $C_{[\cdot]}$. Specifically, for an extended domain output-input pair $(X_{[L]}, Y_{[L]})$ of size $L$, we define the composite denoiser $D_{[L]}$ as
\begin{flalign}%
    \label{eq:dde_denoiser}
    &D_{[L]}(X_{[L]}(t), Y_{[L]}, t) = \nonumber \\
    & \qquad \qquad \quad
    C_{[L]}\Big(
    [D(x_i(t), y_i, t)]_{i=1}^L, [y_i]_{i=1}^L, t\Big),
\end{flalign}%
where $([x_i(t)]_{i=1}^L, [y_i]_{i=1}^L) = F(X_{[L]}(t), Y_{[L]})$. The coordinator network $C_{[\cdot]}$ takes two sequences as inputs: 1) the outputs $[D(x_i(t), y_i, t)]_{i=1}^L$ of the pre-trained denoiser $D$ evaluated on the smaller objects; 2) the conditioning information $[y_i]_{i=1}^L$, and produces an estimate of $X_{[L]}(0)$. 

We train the coordinator by minimizing the denoising error \eqref{eq:denoising} of the composite denoiser \eqref{eq:dde_denoiser} on a training dataset $\mathcal{D}^N_{[L_\text{train}]}$ of objects of size $L_\text{train}$.

\begin{align}
    \label{eq:denoising_L_train}
    &\mathcal{L}_{[L_\text{train}]} = \E_{t \sim p(t)} \E_{\substack{(X_{[L_\text{train}]}, Y_{[L_\text{train}]})  \sim  \mathcal{D}^N_{[L_\text{train}]}}
    } \E_{\varepsilon \sim \mathcal{N}(0; t^2)} \Big[ \nonumber\\
    & \quad
        \lambda(t) \big\|D_{[L_\text{train}]}(X_{[L_\text{train}]} + \varepsilon, Y_{[L_\text{train}]}, t) - X_{[L_\text{train}]}\big\|_2^2
    \Big].
\end{align}
After the coordinator $C_{[\cdot]}$ has been trained, we use it to generate objects of sizes $L_\text{test} \geq L_\text{train}$ to evaluate the generalization to the generation of larger objects.

\subsection{Coordinator Architecture}
\label{ssec:architecture}

For each of our experimental domains, we show that a visual transformer is a suitable choice for the architecture of the coordinator $C$, allowing for better generalization, faster training, and a lower number of parameters compared to the base model. We choose to use the architecture from \cite{peebles2023scalable}, adapting it to the particular domains. For positional encodings of tokens for larger object generation, we used Rotary Position Embedding~\cite{su2023roformer}.

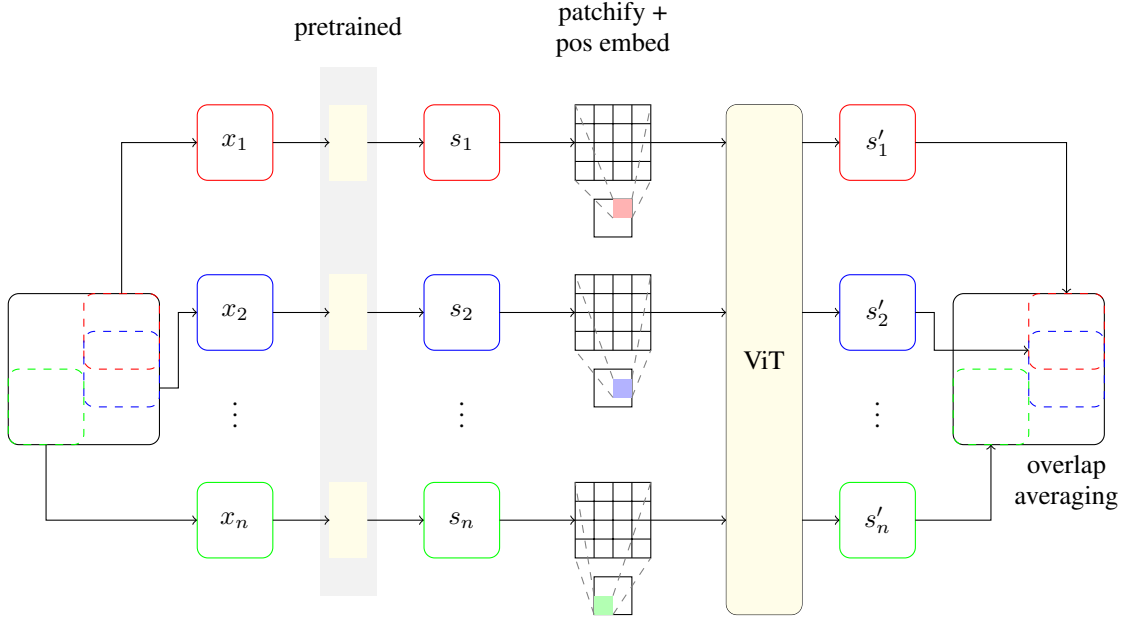
\begin{figure*}[h!]
\centering
\begin{tikzpicture}[scale=0.5]
    \draw[rounded corners] (-2,0) rectangle (2,4);
    \draw[rounded corners, draw=red, dashed] (0,2) rectangle (2,4);
    \draw[rounded corners, draw=blue, dashed] (0,1) rectangle (2,3);
    \draw[rounded corners, draw=green, dashed] (-2,0) rectangle (0,2);
    \draw[->] (1,4) -- (1,8) -- (3,8);
    \draw[->] (2,1.5) -- (2.5,1.5) -- (2.5,3.5) -- (3,3.5);
    \draw[->] (-1,0) -- (-1,-2) -- (3,-2);
    
    \draw[rounded corners, draw=red] (3,7) rectangle (5,9);
    \node at (4,8) {$x_1$};
    \draw[rounded corners, draw=blue] (3,2.5) rectangle (5,4.5);
    \node at (4,3.5) {$x_2$};
    \draw[rounded corners, draw=green] (3,-3) rectangle (5,-1);
    \node at (4,-2) {$x_n$};
    \node at (4,1) {$\vdots$};

    \fill[gray!10] (6.25,-4) rectangle (7.75,10);
    \fill[yellow!10] (6.5,7) rectangle (7.5,9);
    \fill[yellow!10] (6.5,2.5) rectangle (7.5,4.5);
    \fill[yellow!10] (6.5,-3) rectangle (7.5,-1);
    \node at (7,11) {pretrained};
    \draw[->] (5,8) -- (6.5, 8);
    \draw[->] (5,3.5) -- (6.5, 3.5);
    \draw[->] (5,-2) -- (6.5, -2);

    \draw[->] (7.5,8) -- (9, 8);
    \draw[->] (7.5,3.5) -- (9, 3.5);
    \draw[->] (7.5,-2) -- (9, -2);

    \draw[rounded corners, draw=red] (9,7) rectangle (11,9);
    \node at (10,8) {$s_1$};
    \draw[rounded corners, draw=blue] (9,2.5) rectangle (11,4.5);
    \node at (10,3.5) {$s_2$};
    \draw[rounded corners, draw=green] (9,-3) rectangle (11,-1);
    \node at (10,-2) {$s_n$};
    \node at (10,1) {$\vdots$};

    \draw[->] (11,8) -- (13, 8);
    \draw[->] (11,3.5) -- (13, 3.5);
    \draw[->] (11,-2) -- (13, -2);

    \node at (14,11) [align=center] {patchify + \\ pos embed};

    \draw[step=0.5] (13,7) grid (14.999,8.999);
    \draw (13,7) rectangle (15,9);
    \draw (13.5,5.5) rectangle (14.5,6.5);
    \fill[red!30] (14,6) rectangle (14.5,6.5);
    \draw[gray, dashed] (14.5,6.5) -- (15,9);
    \draw[gray, dashed] (14.5,6) -- (15,7);
    \draw[gray, dashed] (14,6.5) -- (13,9);
    \draw[gray, dashed] (14,6) -- (13,7);

    \draw[step=0.5] (13,2.5) grid (14.999,4.499);
    \draw (13,2.5) rectangle (15,4.5);
    \draw (13.5,1) rectangle (14.5,2);
    \fill[blue!30] (14,1.25) rectangle (14.5,1.75);
    \draw[gray, dashed] (14.5,1.75) -- (15,4.5);
    \draw[gray, dashed] (14.5,1.25) -- (15,2.5);
    \draw[gray, dashed] (14,1.75) -- (13,4.5);
    \draw[gray, dashed] (14,1.25) -- (13,2.5);

    \draw[step=0.5] (13,-2.999) grid (14.999,-1.001);
    \draw (13,-3) rectangle (15,-1);
    \draw (13.5,-4.5) rectangle (14.5,-3.5);
    \fill[green!30] (13.5,-4.5) rectangle (14,-4);
    \draw[gray, dashed] (14,-4) -- (15,-1);
    \draw[gray, dashed] (14,-4.5) -- (15,-3);
    \draw[gray, dashed] (13.5,-4) -- (13,-1);
    \draw[gray, dashed] (13.5,-4.5) -- (13,-3);

    \draw[->] (15,8) -- (17, 8);
    \draw[->] (15,3.5) -- (17, 3.5);
    \draw[->] (15,-2) -- (17, -2);

    \draw[rounded corners] (17,-4.5) rectangle (19,9);
    \fill[yellow!10, rounded corners] (17,-4.5) rectangle (19,9);
    \node at (18,2.25) {ViT};

    \draw[->] (19,8) -- (20, 8);
    \draw[->] (19,3.5) -- (20, 3.5);
    \draw[->] (19,-2) -- (20, -2);

    \draw[rounded corners, draw=red] (20,7) rectangle (22,9);
    \node at (21,8) {$s_1'$};
    \draw[rounded corners, draw=blue] (20,2.5) rectangle (22,4.5);
    \node at (21,3.5) {$s_2'$};
    \draw[rounded corners, draw=green] (20,-3) rectangle (22,-1);
    \node at (21,-2) {$s_n'$};
    \node at (21,1) {$\vdots$};

    \draw[->] (22,8) -- (26, 8) -- (26, 4);
    \draw[->] (22,3.5) -- (22.5,3.5) -- (22.5,2.5) -- (25, 2.5);
    \draw[->] (22,-2) -- (24,-2) -- (24, 0);

    \draw[rounded corners] (23,0) rectangle (27,4);
    \draw[rounded corners, draw=red, dashed] (25,2) rectangle (27,4);
    \draw[rounded corners, draw=blue, dashed] (25,1) rectangle (27,3);
    \draw[rounded corners, draw=green, dashed] (23,0) rectangle (25,2);
    \node at (26,-1) [align=center] {overlap \\ averaging};
\end{tikzpicture}
\caption{Overview of the architecture. A large image is decomposed into a set of overlapping patches; each patch is processed by a separate pre-trained denoising network. The denoised outputs of pre-trained models are patchified and augmented with global positional encoding. The coordinator processes all patches and produces a new coordinated set of output patches. The final expanded denoised output is constructed from coordinator outputs by averaging the values in the overlaps.}
\label{fig:arch}
\end{figure*}

To generate larger objects, we cover them with overlapping patches of a size equal to the generation capability of the base model and pass down the predicted denoised versions of the patches during training into the ViT. Our transformer architecture encodes the positional encoding of the resulting tokens relative to the larger object, and during the forward process, we reconcile the predicted overlapped patches with their mean value for each position, inspired by the sampling method from MultiDiffusion \cite{bar2023multidiffusion}. The architecture of the coordinator can be seen in Figure \ref{fig:arch}.

\begin{figure}[H]
        \centering
        \captionsetup{type=table} 
        \caption{FAD values for long music track generation on Slakh2100 task for different architectures. The base model was trained for $80$ epochs and has $405M$ parameters.}
        \label{tab:eval}
        \resizebox{\linewidth}{!}{%
            \begin{tabular}{llll}
                \toprule
                Method&  Size& FAD for $4l$ & FAD for $10l$ \\
                \midrule
                Concat & -  & 4.623 & 4.596 \\
                MultiDiffusion & - & 4.732 & 4.796   \\
                RNN      & $16M$ & 4.223 & 4.081  \\
                RNN with overlaps& $50M$ & 4.447 & 4.424 \\
                DDE (ours) & $66M$ & \textbf{2.112} & \textbf{2.142} \\
                \bottomrule
            \end{tabular}
        }
        \captionsetup{type=table} 
        \caption{Accuracy on the 256 generated samples for various methods and numbers of conditionings on CLEVER conditional image generation.} 
        \label{tab:accuracy}
        \resizebox{\linewidth}{!}{%
            \begin{tabular}{lllllll}
                \toprule 
                \textbf{Model}&  \textbf{Sampler}& \multicolumn{5}{c}{\textbf{Coordination}}  \\
                &   & 1 & 2 &  3 & 4  & 5  \\
                  \midrule
                 RRR & Euler & 98.0 & 93.8 & 72.3 & 48.0 & 23.0 \\
                 & Heun &  \textbf{98.0} & \textbf{94.5} & 83.2 & 63.3 &  33.6 \\
                \midrule
                 MultiDiffusion & Euler & 96.9 & 94.1 & 76.2 & 43.0 & 25.0  \\
                  & Heun & 97.7 & 93.4 & 80.5 & 58.6 & 36.3  \\
                \midrule
                DDE (ours) & Heun & 96.5 & 94.1 & \textbf{86.3} & \textbf{66.8} & \textbf{44.5}  
                \\
                \bottomrule
            \end{tabular}
        }
        \captionsetup{type=table}
        \caption{Evaluation results for satellite image generation.}
        \label{sample-table}
        \resizebox{\linewidth}{!}{
            \begin{tabular}{lll}
                \toprule
                Method&  FID, $N=96$& FID, $N=128$ \\
                \midrule
                MultiDiffusion & 37.815& 35.016  \\
                DDE (ours) & \textbf{31.753} & \textbf{27.373} 
                \\
                \bottomrule
            \end{tabular} 
        }
    \vskip-2em
\end{figure}

\begin{figure}[t]
        \centering        \includegraphics[width=0.45\linewidth]{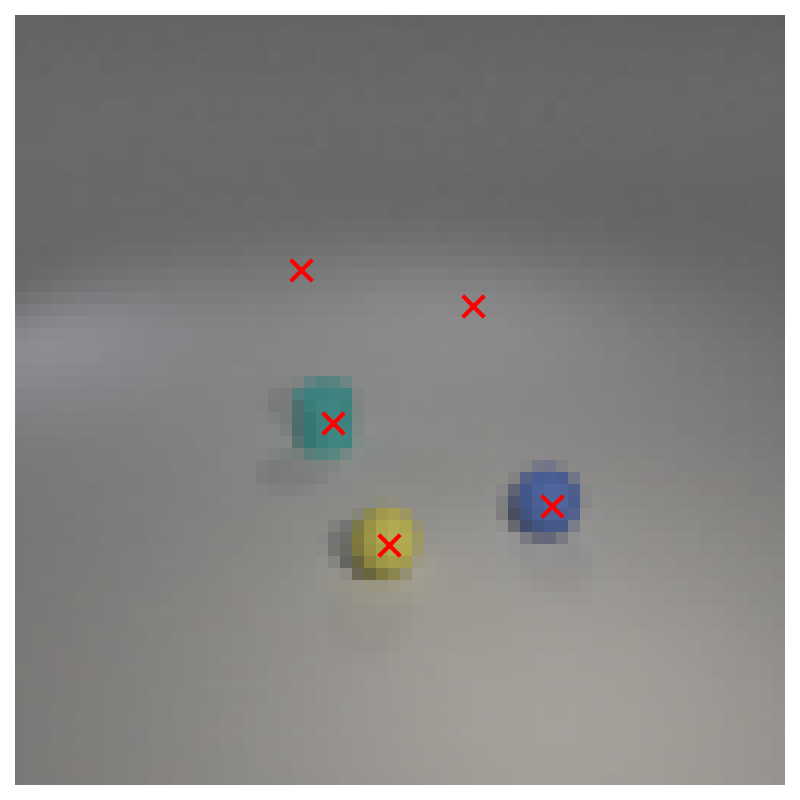}
         \includegraphics[width=0.45\linewidth]{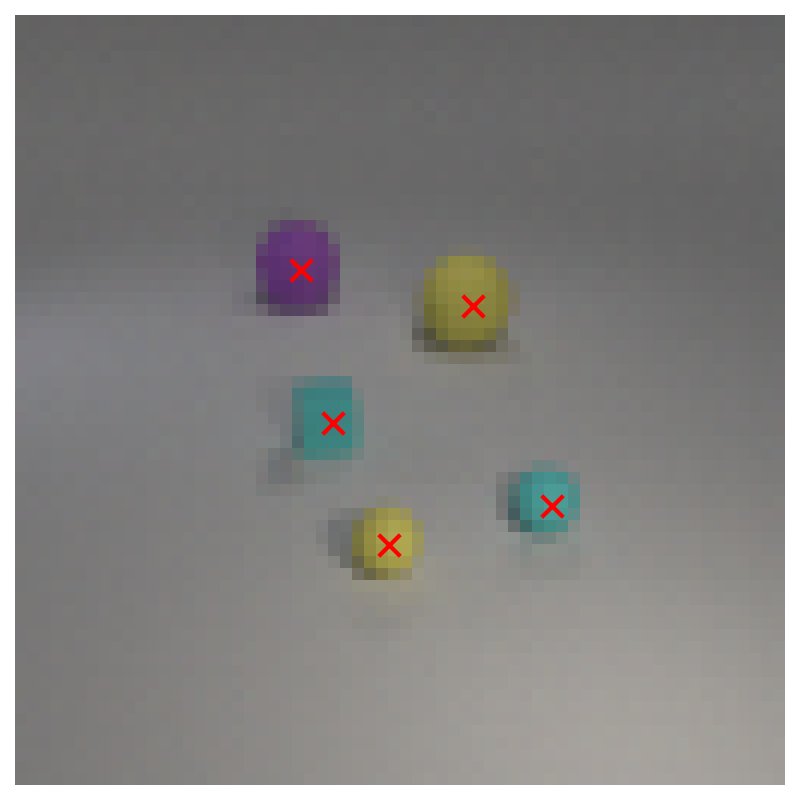}\\
        \captionsetup{type=figure} 
        \caption{Generated sample from the RRR model (left) and Coordinator model (right). Red crosses correspond to the conditionings.}
        \label{fig:sample}\includegraphics[trim={11cm 0 0 0},clip,width=0.80\linewidth]{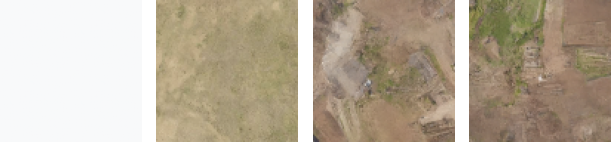}
        \captionsetup{type=figure}
        \caption{Satellite image generation example: DDE sample on the left and MultiDiffusion sample on the right.}
        \label{fig:mdiff96}
    \vspace{0em}
\end{figure}

\section{Experiments}
\label{sec:experiments}
We set up the experiments to highlight the ability of a coordinator model trained on the denoised outputs to perform diverse tasks. In music generation, we pretrain models that can generate short tracks and then generate larger tracks by coordinating these models. In the CLEVR dataset, we perform conditioned sampling with multiple conditionings by coordinating several models that are capable of handling only one conditioning. Finally, in the satellite image generation domain, we combine both the idea of generating a larger object and applying conditionings.

\subsection{Music Generation}
\label{ssec:exp_music}

First, we pre-trained a base model with UNet architecture on Slakh2100 \citep{manilow2019cutting} capable of generating music tracks of length $l$, which translates into $12$ seconds of music. In this experiment, we chose $L_{\text{train}} = 5$ and $L_{\text{test}} = 13$. We evaluate the method by computing FAD \citep{kilgour2019frechet} between the generated long tracks and the sampled tracks of the same length from the dataset.

Our baselines are concatenation and MultiDiffusion \citep{bar2023multidiffusion}. The concatenation method applies a pre-trained model on parts of the large track independently and concatenates the results during inference. The MultiDiffusion method decomposes a large track into a sequence of overlapping patches. In each step of the diffusion process, the pre-trained model is run on each of the patches, and the score at each position is the mean of all scores from the patches that contain this position.

We report the FAD metric based on $128$ generated samples for various architectures, and $48$ and $120$ seconds of music generation (Table \ref{tab:eval}). We describe all the architectures we trained in detail in Appendix \ref{ssec:app_clever_details}. To train and test RNN with overlaps and DDE, for lengths of $4l$ and $10l$, we select $5$ and $13$ patches of length $l$ accordingly, so that the overlap of two adjacent patches has a length of $\frac{l}{4}$. We note that, despite not having access to tracks of length $10l$ during training, the coordinator architectures were able to generalize and generate coherent long tracks while performing better than both concatenation and MultiDiffusion. One property we observe is that the coordinator models are much smaller than the base model, and their training converges faster. ViT, which we used in DDE, performs significantly better in terms of the metric on the tracks of both lengths, compared to other methods.


\subsection{CLEVR Image Generation}
\label{ssec:exp_clevr}

To test the ability of DDE to expand the number of conditioning inputs, we conducted experiments with the conditional generation problem used in \citep{du2023reduce} on the CLEVR dataset \cite{johnson2016clevr}. In this problem, the pre-trained model is a UNet-based conditional diffusion model $p(x \vert c)$, where $x$ is a $64 \times 64$ image and $c$ is conditioning information consisting of a pair of Euclidean coordinates $(c^x, c^y)$. The condition corresponds to the constraint of the image having an object centered at position $(c^x, c^y)$, in addition to possibly having other objects located elsewhere. The goal of the problem is to generate images $x \sim p(x \vert [c_i]_{i=1}^L)$ conditioned on objects being present at positions $[c_i]_{i=1}^L$. In the notation of Section~\ref{ssec:method_general}, the expanded generative domain in this problem is $Y_{[L]} = [c_1, \ldots, c_L]$, $X_{[L]} = [x, \ldots, x]$: generating one image $x$ conditioned on multiple positions $[c_i]_{i=1}^L$. We evaluate the generated samples $x$ using a classifier $\operatorname{CLS}(x, c)$, which, given an image $x$ and a position $c$, outputs a probability of an object being present in the image at position $c$. Following \citet{du2023reduce}, we trained base diffusion models that realize conditional, unconditional, and classifier-free guided sampling \citep{ho2021classifierfree} with a shared parameterization. We trained a ViT-based coordinator $C_{[\cdot]}$. We provide the details about base model training in Appendix \ref{ssec:app_clever_base_model}, and about the coordinator architecture and training in Appendix \ref{ssec:app_clever_coord}.

During training, the coordinator observed conditioning objects of size up to $L_\text{train} = 2$. For evaluation, we tested conditional generation with conditioning objects of sizes up to $L_\text{test} = 5$, and calculated the overall accuracy for each number of conditionings; see Appendix \ref{ssec:app_clever_eval} for details.

We considered two baseline methods. For the first baseline, we considered the approach developed in \cite{du2023reduce} (denoted ``RRR''), which combines conditioned and unconditioned scores of the pre-trained model on several conditionings. We provide the concrete formula in Appendix \ref{ssec:app_clever_eval}. As another baseline, we evaluated how the method from MultiDiffusion works by simply averaging the outputs of the pre-trained model over all conditionings. Figure~\ref{fig:sample} shows samples from the RRR model and our Coordinator model.


Table \ref{tab:accuracy} shows the conditional generation accuracy of our method and the baselines with various sampling methods. Our coordinator outperforms the baselines significantly, especially when conditioned on $4$ or $5$ labels. The results demonstrate that the transformer architecture can generalize more conditioning inputs than observed during training. Our architecture also uses slightly fewer parameters than the base model ($38M$ versus $42M$) and uses significantly fewer epochs of training ($20$ versus $1050$).


\subsection{Map Image Generation}
\label{ssec:exp_maps}

Our final experimental domain is conditional image generation. We chose the task of generating satellite images conditioned on schematic maps. We collected a new dataset for this task using the Google Maps API.

We pretrain a model that can sample patches of size $B \times B$ where $B = 64$. Our decomposition function selects $B \times B$-size patches from a larger image with a stride $s = 32$. We train the coordinator on size $N \times N$, where $N = 96$, and perform evaluation for out-of-distribution sampling with $N = 128$. We use FID between the generated conditional samples and the set of all ground truth satellite images as the evaluation metric. The baseline with which we are comparing our method is MultiDiffusion.

The base model is trained for 200 epochs and has 270M parameters. The coordinator has 26M parameters and is trained for 20 epochs. Table~\ref{sample-table} displays the value of the FID metric for both the training image size ($N = 96$) and the out-of-distribution sampling size ($N = 128$). Our method outperforms MultiDiffusion by FID.

In Figure \ref{fig:mdiff96}, we present samples from DDE and multidiffusion. We note that top-right part of the multidiffusion sample that lies within only one patch is visually separable from the rest of the image. In Appendix \ref{sssec:map_samples}, we show more DDE samples.





\section{Conclusion}
\label{sec:conclusion}

In this paper, we proposed Diffusion Domain Expansion (DDE), a method that utilizes a ViT network as a coordinator of outputs from pre-trained diffusion models to expand the generative domain. We demonstrated that our method allows for the generation of larger objects and the combination of multiple conditioning inputs simultaneously. Our experiments indicate that DDE enables the coordinator to generalize to larger instances unseen during training. Lastly, our model requires less training time and fewer parameters to converge to the optimal state compared to the base pre-trained models.

\textbf{Limitations}. The main limitation of our method is the need for additional data and time to train the coordinator model.

\textbf{Future work}. We plan to explore the application of our method in other domains. Potential areas include generating long coherent protein sequences and creating videos from short clips. In addition, we are interested in exploring alternative ways of supervision for the coordinator model.


\bibliography{refs}
\bibliographystyle{icml2024}

\newpage

\appendix
\onecolumn

\counterwithin{figure}{section}
\counterwithin{table}{section}
\counterwithin{algorithm}{section}

\section{Related Work}
\label{sec:related}

\paragraph{Controllable generation.}
Generation control and guidance of diffusion models is an active area of research. One way to formalize controllable generation is to cast it as sampling from a conditional distribution $p(x \vert y)$. Popular types of conditioning annotations $y$ include class labels \citep{dhariwal2021diffusion}, text prompts \citep{ramesh2021zero, rombach2022high, saharia2022photorealistic}, and semantic maps \citep{rombach2022high, huang2023collaborative}. Method-wise, a straightforward way to achieve conditional generation $p(x \vert y)$ is to train a conditional denoising network $D(x, y, t)$ on pairs $(x, y)$ of generation target $x$ and conditioning information $y$ available at training time. Classifier-free guidance \citep{ho2021classifierfree} enhances image fidelity by combining conditional and unconditional score functions.

Another family of methods enables generation conditioned on information that was not specified at model training time. Classifier guidance \citep{sohl2015deep, dhariwal2021diffusion, song2021scorebased} generates samples by combining the learned score function and the gradient of the log-probabilities of a separately trained classifier $p(y \vert x, t)$. \citet{zhang2023adding} developed ControlNet, a method that employs a supplementary trained network to enable a pre-trained diffusion model to process previously inaccessible conditioning inputs. Collaborative Diffusion \citep{huang2023collaborative} trains a lightweight module which combines scores of multiple diffusion models, each supporting different conditioning modalities, to enable multi-modality conditioning.

Image inpainting is an extensively explored controllable generation task. The goal of inpainting is to restore or reconstruct missing or corrupted parts of an image, or, formally, sampling $x_\text{hidden} \vert x_\text{obs} \sim p(x_\text{hidden} \vert x_\text{obs})$, the unobserved part of the image $x_\text{hidden}$ given the observed part $x_\text{obs}$. \citet{song2021scorebased, lugmayr2022repaint} proposed inpainting techniques based on the "replacement" of observed pixels with their known values (with noise applied at an appropriate) at each step of the diffusion denoising process. \citet{chung2022improving, ho2022video} utilized "reconstruction"-based approaches which introduce an additional guiding term in the denoising update with the goal of bringing the values in the observed part of the image closer to the known target values. \citet{trippe2023diffusion} developed a method based on particle filtering for inpainting of protein backbones in 3D. \citet{saharia2022palette} proposed a diffusion model trained specifically for image-to-image translation tasks including inpainting.

The task of inpainting is a member of the family of inverse problems, which can be cast as conditional generation $x \vert y(x) \sim p(x \vert y(x))$, where $y(x)$ is an observation function that extracts a limited information summary from $x$ (e.g., $y(x)$ might represent a downsampled version of image $x$). A line of work \cite{choi2021ilvr, song2021scorebased, kawar2022denoising, saharia2022palette, whang2022deblurring, ren2023multiscale} is focused on diffusion-based solutions for inverse problems in the image generation domain. Recently, \citet{mariani2024multisource} used diffusion models for the inverse problem of audio source separation. \citet{ben2024d} utilized the probability flow interpretation of diffusion models and addressed image and audio inverse problems and conditional molecular generation via differentiation through ODE sampler.

Image editing is another instance of controllable generation problems. In this case, the goal is to modify specific aspects of an existing image according to user-specified goals while preserving image coherence and fidelity. To address semantic image editing, \mbox{\citet{meng2022sdedit}} and \citet{couairon2023diffedit} proposed to first partially noise and then denoise an image to generate an edited version, possibly conditioned on a segmentation mask~\citep{couairon2023diffedit}. Collage diffusion \citep{sarukkai2024collage} is a diffusion model extension that processes multiple base images accompanied by text prompts and desired locations and produces a collage by editing the base images and combining layers of edited images. Another line of work \cite{ruiz2023dreambooth, gal2023an, hertz2023prompttoprompt, kawar2023imagic} introduced techniques for personalization, concept learning, prompt manipulation, and direct editing of real images guided by natural language inputs.

Fine-tuning diffusion models with reward functions or human preference data has emerged as a promising form of controllable generation. \citet{black2024training} interpreted diffusion as a sequential decision process and employed reinforcement learning to optimize objectives such as image compressibility, prompt-image alignment (based on vision-language model feedback), and aesthetic quality (derived from human feedback). \citet{clark2024directly} used backpropagation through samplers to optimize differentiable rewards such as scores from human preference models. \citet{wallace2023diffusion} adapted Direct Preference Optimization \citep{rafailov2023direct} to diffusion models, which enables fine-tuning diffusion models on human preference data directly without the need for the auxiliary reward model.

\paragraph{Diffusion composition and coordination.}
Compositional generation approaches combine multiple diffusion processes to control sampling distributions, re-use pre-trained models, and extend their capabilities.
Existing works on compositional generation differ in how they approach the composition of diffusion processes. Notable approaches include exact specification of adjusted distributions \cite{du2023reduce, garipov2023compositional, wu2024compositional}, or resolution of individual steps of generation \cite{bar2023multidiffusion, lee2023syncdiffusion, huang2023collaborative, zhang2023diffcollage, corso2024particle}.

\citet{garipov2023compositional} proposed a set of operations on iterative generative processes: diffusion models and GFlowNets \citep{bengio2021flow}. These operations enable one to emphasize or de-emphasize high-likelihood regions of selected base models. The proposed Compositional Sculpting method employs classifier guidance and mixture processes to compose models and realize samples from composite distributions.

\citet{wu2024compositional} built on compositions of energy-based diffusion models \cite{du2020compositional, du2023reduce} and developed methods for the generation of solutions to inverse design problems: multi-body physical simulations and 2D airfoil design. The compositional generation enables the generation of objects more complex than those seen at training time. In particular, the authors demonstrate the generation of trajectories with 1) longer simulation horizons (via summation of energies of short-horizon models), 2) larger number of interacting bodies in physical simulations (via summation of energy functions for pairwise interactions), and 3) larger spatial simulation domains and the larger number of parts for joint multiple airfoil design.

MultiDiffusion \citep{bar2023multidiffusion} performs controlled image generation via the fusion of diffusion sampling paths. The authors address the generation of panoramic images and scenes with complex structures. The complex scene is divided into several regions, and the desired content of each region is described with a specific text prompt. The generation is performed via the coordination of multiple prompt-conditioned generation paths. The coordination method maintains a shared global image for the whole space and reconciles contradicting updates in the intersections of regions by averaging the updates of individual models whose regions cover the specific intersection.

SyncDiffusion \citep{lee2023syncdiffusion} is an extension of MultiDiffusion that aims to address the issue of incoherent patches in large panoramic images. To that end, SyncDiffusion introduces an additional step to the MultiDiffusion update to perform a local optimization step on the perceptual similarity across patches.

\citet{zhang2023diffcollage} extended panoramic generation to more general scenarios of large image generation supporting arbitrary graph structure of the overlapping patches comprising the large image (e.g., linear chain, cycle, grid, cubemap). The proposed method, called DiffCollage, translates a given graph structure into a closed-form formula for the total score expressed through marginal scores of individual patches. \citet{zhang2023diffcollage} empirically demonstrated that coordinating multiple diffusion processes run in parallel outperforms inpainting-based panoramic image generation \cite{lugmayr2022repaint, chung2022improving} in terms of image quality and generation speed.

\citet{corso2024particle} proposed Particle Guidance, a method for improving sample efficiency in diffusion models by running multiple diffusion chains with a time-evolving repulsion force that promotes sample diversity.

\citet{mariani2024multisource} demonstrated that the combination of multiple single-instrument diffusion models outperforms a joint model in the music source separation problem. The proposed source separation approach is based on the guidance methods for inverse problems proposed in \cite{song2021scorebased}.

\section{Experiment Details}
\label{sec:app_exp_details}

All training for both the base and coordinator models was performed using the EDM framework introduced by \cite{karras2022elucidating} on a single A100 GPU. To enhance the quality of the generated samples, we trained the coordinator using EMA for the model weights. As shown by \cite{izmailov2019averaging} and \cite{karras2022elucidating}, this approach improves generalization and made training more stable in terms of metric values per epoch during evaluation.

For sampling from our coordinator models, we primarily used the Heun sampler with 2nd order correction, which we imported from the code provided by \cite{karras2022elucidating}.

\subsection{Music Generation Experiments}
\label{ssec:app_music_details}

\subsubsection{Dataset description}
\label{ssec:app_music_dataset}
For the music domain, we used the version of the Slakh2100 dataset released by \citet{mariani2024multisource} for training both the base model and coordinator models. This dataset contains 1,500 tracks in the training set, each consisting of four different stems: Bass, Drums, Guitar, and Piano.

\subsubsection{Base model details}
\label{ssec:app_music_base_model}

For training the base model, we used the same UNet1d architecture and data loader as in \cite{mariani2024multisource}. The hyperparameters were set as follows: $channels=256$, $patch\_factor=16$, $patch\_blocks=1$, $resnet\_groups=8$, $kernel\_sizes\_init=[1, 3, 7]$, $multipliers=[1, 2, 4, 4, 4, 4, 4]$, $factors=[4, 4, 4, 2, 2, 2]$, $num\_blocks=[2, 2, 2, 2, 2, 2]$, $attentions=[False, False, False, True, True, True]$, $attention\_heads=8$, $attention\_features=128$, and $attention\_multiplier=2$. This architecture takes as input four different stems of length $2^{18}$, corresponding to roughly 12 seconds of music at a sampling rate of 22,050 Hz, and produces output of the same size.

We used the Adam optimizer with a learning rate of $10^{-4}$, with $betas = (0.9, 0.99)$, and trained the model for 300 epochs with a batch size of 8. We sampled the noise level $\sigma$ from a log-normal distribution with a mean of $-3$ and a standard deviation of $1.0$, and we did the same for the training of coordinator models.

\subsubsection{Sampler details}
\label{ssec:app_music_sampler}

To sample the tracks for all the methods and architectures, we used the Heun sampler described in Appendix \ref{sec:app_exp_details} with $\sigma_{\max} = 20$, $\sigma_{\min} = 10^{-4}$, and $S_{\text{churn}} = 20.0$. For $\sigma_t$, we used the Karras schedule with $\rho = 7$ and 150 timesteps.

\subsubsection{Evaluation}
\label{ssec:app_music_eval}

To evaluate the performance of all the models, we calculated the FAD (Fréchet Audio Distance) to the training dataset. To calculate the FAD metric for the sampled tracks of length $l$, we deterministically cut out the middle parts of the tracks from the training dataset with the same length, as the FAD metric is track-length dependent. In the implementation of the metric calculation we used, a pre-trained VGG-like architecture for music was used. Since the sampling rate was different from the required one, we resampled tracks from 22,050 Hz to 16,000 Hz. We tested how different numbers of sampled tracks affect the FAD score and found that generating 128 tracks differs from generating 1,024 tracks by a small margin, and the relative order of the compared models does not change. To decrease GPU usage, we settled on using 128 tracks in our reported metric values. We reported the saved checkpoint with the smallest FAD value for each of the methods.

\subsubsection{RNN and RNN with overlaps details}
\label{ssec:app_music_rnn}
We explored different methods of coordinating the denoised tracks via trainable networks. The first method we propose is a combination of RNN and UNet architectures. We choose a constant $h$—the number of hidden channels—and train a function $f\colon \mathbb{R}^{h \times L} \times \mathbb{R}^{S \times L} \to \mathbb{R}^{h \times L} \times \mathbb{R}^{S \times L}$ that is parametrized as a UNet with $h + S$ input channels and $h + S$ output channels. Then, let the predicted denoised tracks before reconciliation be $x_0, \ldots, x_{k-1}$. We set $h_0 = 0$, and $(h_{i+1}, x'_i) = f(h_i, x_i)$ for $i = 0, 1, \ldots, k-1$. The output of the RNN denoiser is the concatenation of $x'_i$, $i=0, 1, \ldots, k - 1$. 

What we defined as RNN with overlaps corresponds to training a similar architecture, which takes patches of music tracks with overlaps of 3 seconds and performs a similar reconciliation by averaging the predicted fixed patches, as in MultiDiffusion. For both of the architectures, we used $h = 16$ hidden channels. We noted that a higher number of hidden channels compared to the number of output channels improved the FAD.

We used the Adam optimizer with a learning rate of $10^{-4}$, with $betas = (0.9, 0.99)$, and trained the models for 10 epochs with a batch size of 4. For the training dataset, we used sampled tracks of length 48 seconds from the dataset, which correspond to 4 non-overlapping or 5 overlapping patches. Despite RNN with overlaps having a slightly worse metric compared to RNN, using overlaps improved the quality of tracks by removing the slightly noticeable transition between different patches.

\subsubsection{ViT details}
\label{ssec:app_music_vit}

Finally, we tested our ViT architecture. For music, we used the ViT architecture with $patch\_size = 128$, $hidden\_size = 768$, $depth = 6$, $num\_heads = 6$, and $mlp\_ratio=4.0$.

We used the Adam optimizer with a learning rate of $3 \cdot 10^{-5}$, with $betas = (0.9, 0.99)$, and trained the models for 10 epochs with a batch size of 4. In addition, for ViT, we obtained the best results using EMA. We tried using EMA for the RNN-like architectures, but it did not improve the metric values.

We chose these three architectures for comparison with the baseline because they can generalize easily enough for the generation of longer tracks, and we chose RNN to compare with ViT, as it requires fewer parameters for training.

We include generated samples from ViT, MultiDiffusion and Concat for both $48$ and $120$ seconds music track length in the supplementary materials.

\subsection{CLEVR Image Generation Experiments}
\label{ssec:app_clever_details}

\subsubsection{Dataset description}
\label{ssec:app_clever_dataset}
For the training of the base model, we used the CLEVR dataset released by the authors of \cite{du2023reduce}. This dataset consists of 30,000 images of size 128 by 128 pixels, which we downsampled to 64 by 64 pixels. For each image in the dataset, we had one conditioning information available, corresponding to the coordinates of the center of one of the objects in the image.

\subsubsection{Base model details}
\label{ssec:app_clever_base_model}
For the training of the base model with one conditioning, we used a UNet with attention architecture from \cite{du2023reduce}. We used the following hyperparameters for the UNet: $model\_channels = 128$, $num\_res\_blocks = 2$, $channel\_mult = (1, 2, 2, 2)$, $num\_heads = 4$, and $num\_head\_channels = 64$. We improved the positional encoding of the conditioning by using Gaussian Fourier projection, projecting each of the coordinates into the dimension of $2 \cdot model\_channels$.

We used the Adam optimizer with a learning rate of $10^{-4}$ and trained for 1,050 epochs with a batch size of 32. We used the original dataset to train the model to satisfy one conditioning and used classifier-free guidance to improve the quality of generated samples. We sampled the noise level $\sigma$ from a log-normal distribution with a mean of $-1$ and a standard deviation of $1.6$, and we did the same for the training of coordinator models.

The base network $D(x, c, t)$ can optionally take a masked-out conditioning input $D(x, \varnothing, t)$ corresponding to unconditional generation. During coordinator training, we mask out the conditioning information with a probability of $10\%$. At generation time, we use classifier-free guidance with weight $w$:

\begin{equation}
    \label{eq:coordinator_cfg}
    \begin{split}
        (1 + w) \cdot C_{[L]}([D(x(t), c_i, t)]_{i=1}^L, [c_i]_{i=1}^L, t) - 
        w \cdot C_{[L]}([D(x(t), \varnothing, t)]_{i=1}^L, [\varnothing]_{i=1}^L, t).
    \end{split}
\end{equation}

\subsubsection{Classifier details}
\label{ssec:app_clever_class}

To evaluate the models, we trained a UNet classifier on the original dataset, which outputs the probabilities of each pixel being the center of an object. We modified the dataset with both positive and negative examples: for the conditioning from the dataset, the classifier should output a $1$ probability on the corresponding pixel, and it should output $0$ for a random conditioning.

We used a UNet architecture from \texttt{denoising\_diffusion\_pytorch} (\href{https://github.com/lucidrains/denoising-diffusion-pytorch?tab=readme-ov-file}{url}), with $dim=64$ and $dim\_mults = (1, 2, 4)$. We added an additional convolution layer and used a sigmoid function on the outputs.

We used the Adam optimizer with a learning rate of $10^{-5}$, with $betas = (0.9, 0.99)$ and $eps = 10^{-8}$. We trained it for 50 epochs with a batch size of 32. Figure \ref{fig:classifier} shows an example output of the model and that of the classifier.

\subsubsection{Coordinator model details}
\label{ssec:app_clever_coord}

The inputs to the coordinator are the outputs of the pre-trained diffusion $[D(x, c_i, t)]_{i=1}^L$ and the positions $[c_i]_{i=1}^L$ encoded as one-hot 2D maps. The one-hot 2D maps are appended as additional channels to each diffusion output. The coordinator is trained with the denoising loss \eqref{eq:denoising_L_train}. During training, we sample an image $x$ from the dataset, then sample a number of conditioning positions $L \sim \operatorname{Uniform}[1, L_\text{train}]$, and then extract positions $[c_{i}]_{i=1}^L$ of $L$ objects from the image $x$. Since the version of the dataset used by \citet{du2023reduce} had each image annotated with only one object position, we ran the classifier on each sample from the dataset and outputted the centers of the connected components of pixels where the score of the classifier is at least $0.5$ as the possible conditioning inputs, as they correspond to the centers of the objects in the image.

To train the coordinator model, we used the ViT architecture with $patch\_size = 4$, $hidden\_size = 384$, $depth = 12$, $num\_heads = 6$, and $mlp\_ratio = 4.0$.

We used the Adam optimizer with a learning rate of $10^{-5}$, and trained for 40 epochs with a batch size of 16. During training, we randomly sampled 2 random centers of figures as the conditionings to illustrate the effect of generalization.

\subsubsection{Samplers details}
\label{ssec:app_clever_sampler}

To evaluate the methods, we used the Euler sampler, using the ODE equation for the reverse diffusion process with 100 timesteps, and the Heun sampler described in \ref{sec:app_exp_details} with $\sigma_{\max} = 80$, $\sigma_{\min} = 10^{-4}$, and $S_{\text{churn}} = 20.0$. For the Heun sampler, we set $s_{\text{churn}} = 0$, and used $\sigma_{\min} = 10^{-4}$, $\sigma_{\max} = 80$. For $\sigma_t$, we used the Karras schedule with $\rho = 7$ and 100 timesteps. In each evaluation, we set $w = 20$ for the classifier-free guidance weight. We tested which $w$ works best in terms of accuracy, and report results using the same weight. For sampling using the RRR method, we set $w = 4$, as used in the code released by \citet{du2023reduce}.

\subsubsection{Evaluation details}
\label{ssec:app_clever_eval}

For the RRR baseline, we used formula \ref{eq:rrr}:

\begin{align}
   RRR_{[L]}(x(t), [c_i]_{i=1}^L, t) = D(x(t), \varnothing; t) + w \cdot \sum_{i = 1}^{L} \left( D(x(t), c_i; t) - D(x(t), \varnothing; t) \right)
   \label{eq:rrr}
\end{align}

\begin{figure}[h!]
  \begin{minipage}[b]{0.49\linewidth}
    \centering
    \includegraphics[width=\linewidth]{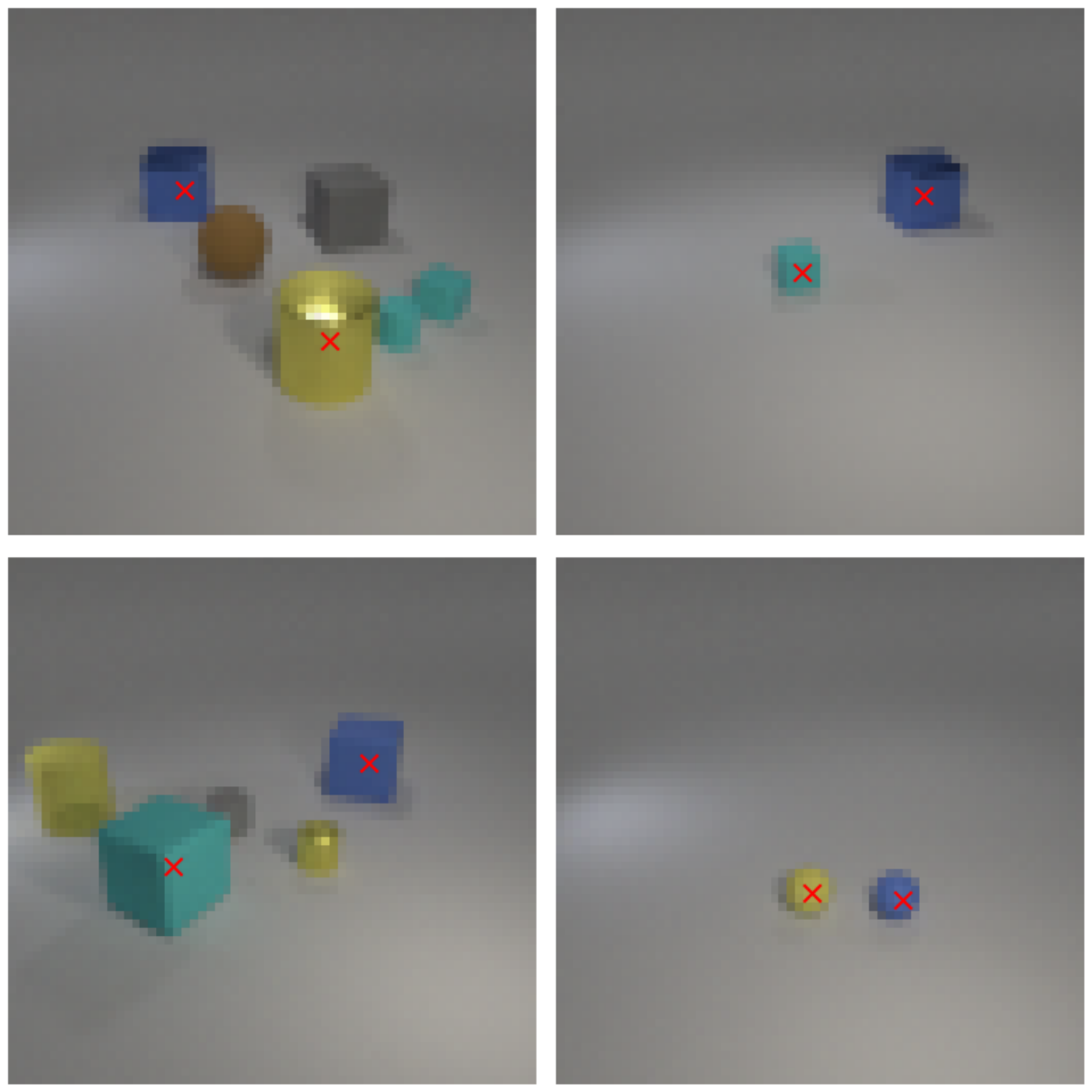}
  \end{minipage}
  \hfill
  \begin{minipage}[b]{0.49\linewidth}
    \centering
    \includegraphics[width=\linewidth]{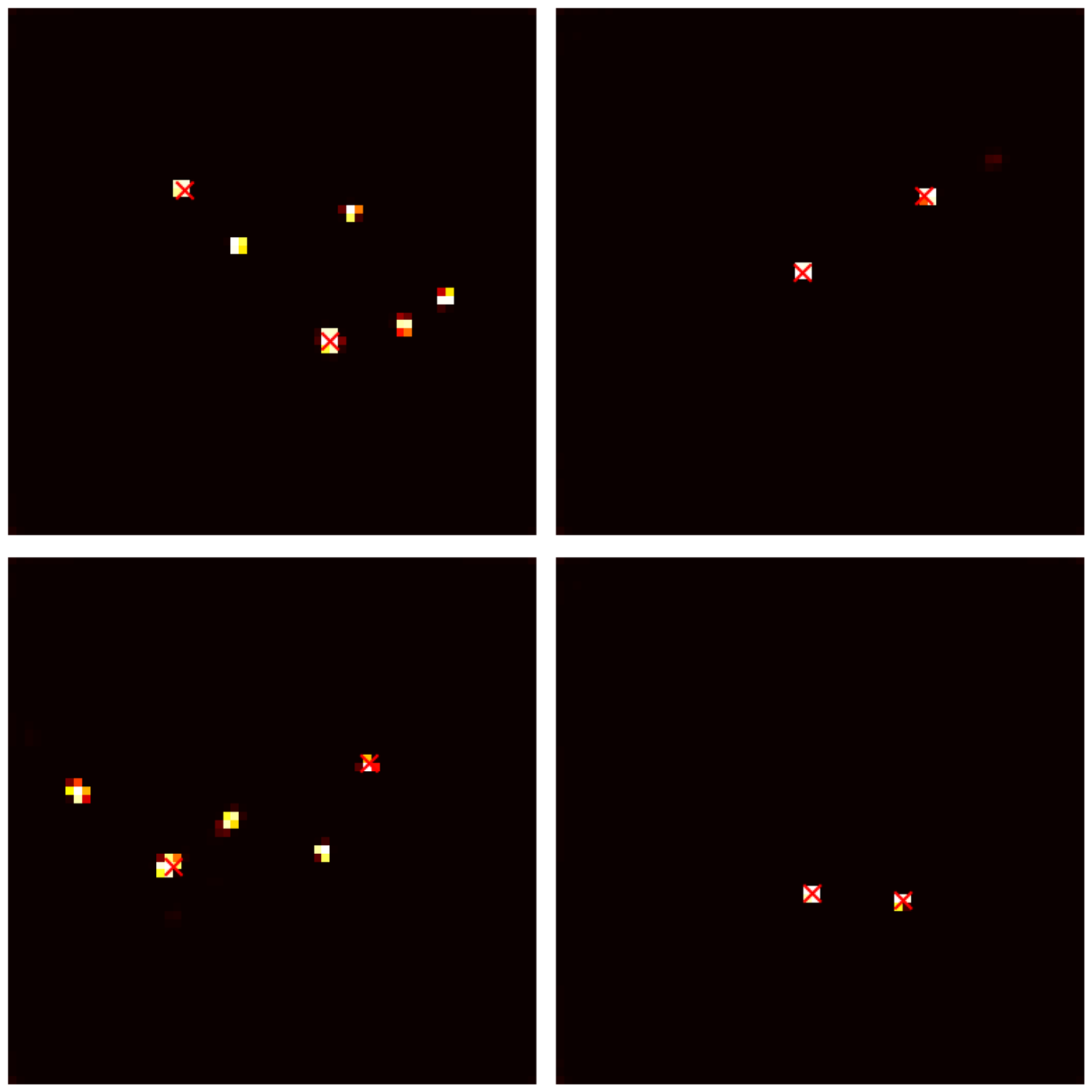}
  \end{minipage}
  \caption{A batch of generated samples from the model based on 2 conditionings and the heatmaps of the classifier output for these samples. Red crosses correspond to the conditionings.}
   \label{fig:classifier}
\end{figure}

For each $L \in \{1, \ldots, L_\text{test}\}$ we generated 256 random sets of positions $[c_i]_{i=1}^L$ and then generated image $x \vert [c_i]_{i=1}^L$ using the pre-trained model and the coordinator. We report the conditional generation accuracy, i.e. the fraction of the generated images satisfying the conditioning constraints. We tested the conditional constraints using the classifier 
$\operatorname{CLS}(x, c)$. Given positions $[c_i]_{i=1}^L$, we consider the generated image $x$ to be valid if $\operatorname{CLS}(x, c_i) \geq 0.5,~\forall\,1 \leq i \leq L$.
We provide an example of the classifier output on a generated sample in \ref{fig:classifier}.

\begin{figure}[h!]
  \begin{minipage}[b]{0.49\linewidth}
    \centering
    \includegraphics[width=\linewidth]{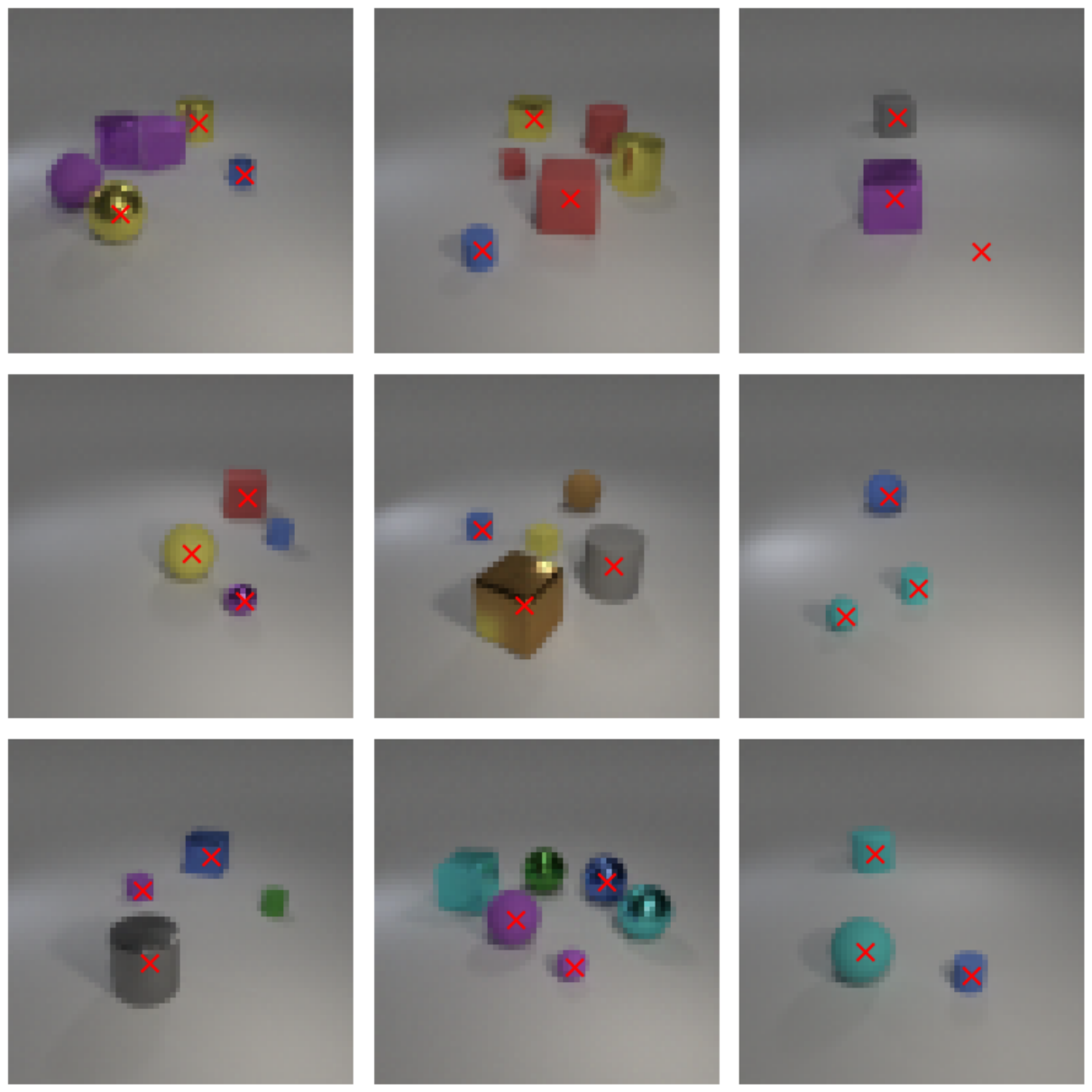}
  \end{minipage}
  \hfill
  \begin{minipage}[b]{0.49\linewidth}
    \centering
    \includegraphics[width=\linewidth]{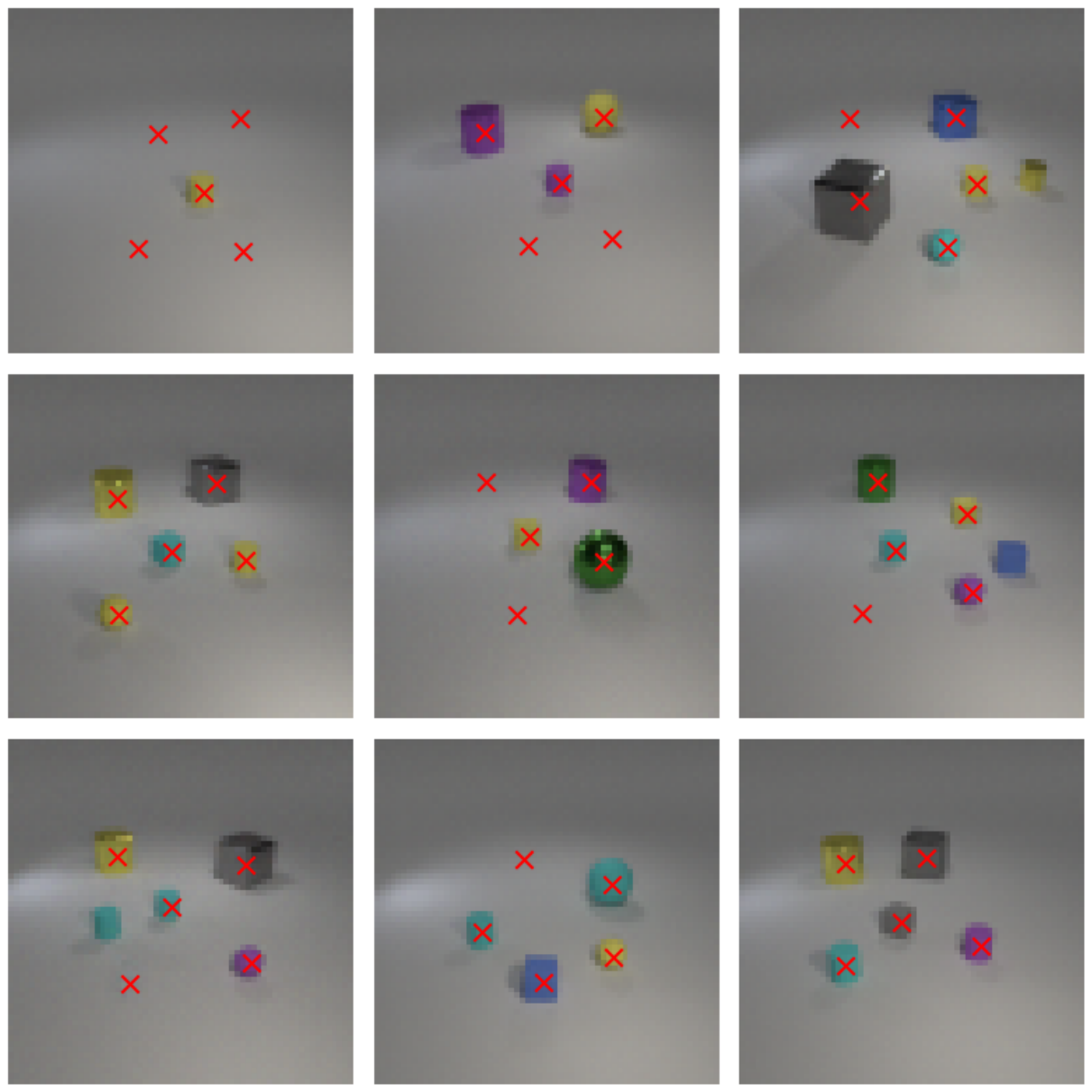}
  \end{minipage}
  \caption{Generated samples from the coordinator model for 3 and 5 different conditionings. Each red cross on the image corresponds to a conditioning.}
  \label{fig:exsamples}
\end{figure}

For each of the 256 samples generated during evaluation, we generated positions $c_1, \dots, c_k$ in the conditioning $Y_{k}$ such that $\forall i: 0.3 \leq c_i^{x}, c_i^{y} \leq 0.7$ and $\forall i, j : ||c_i, c_j||^{2} \geq 0.15$. We imposed these additional constraints on the generation of conditionings to avoid evaluating samples on very close conditioning inputs, as in such cases, a suitable sample could include objects satisfying multiple conditionings simultaneously. We used the same seed for the generation of all samples for each combination of sampler and method to make the comparison fairer. We provide some additional samples generated from our DDE model in \ref{fig:exsamples}. We can see that even when some of the conditionings are not satisfied, our model still tries to satisfy as many of them as possible.

\subsection{Map Image Generation Experiments}
\label{ssec:app_map_details}

\subsubsection{Dataset description}

\begin{figure}[h!]
\centering
\includegraphics[width=0.7\linewidth]{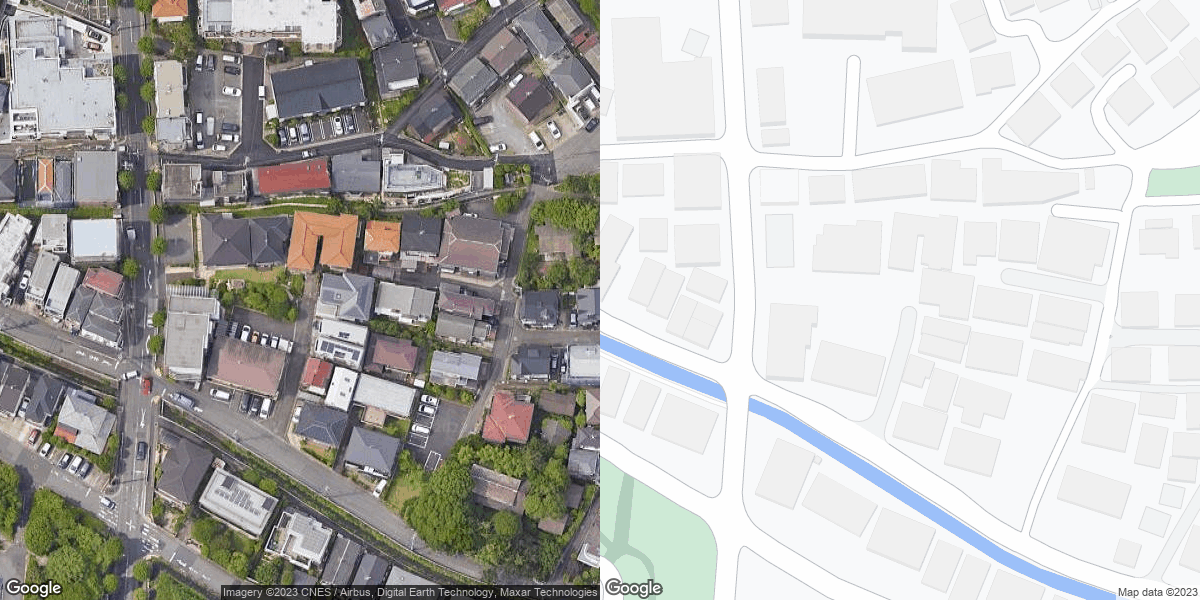}
\caption{Example from the map dataset. Image attribution: Google Maps}
\label{fig:map512}
\end{figure}

We selected 7,300 uniform samples of size $150 \times 150$ meters from a $20 \times 20$ km square, centered at (\(35.707^\circ\) N, \(139.600^\circ\) E). Each sample contains a pair of a map and the corresponding satellite image and has a resolution of $600 \times 600$. We cropped a $512 \times 512$ sub-square from it and downsampled it to $128 \times 128$. Figure \ref{fig:map512} demonstrates a raw dataset example before cropping and downsampling.

\subsubsection{Architecture details}

We use decomposition functions that take small patches inside the map. Specifically, we have the following parameters: the size of the patch $B$, the stride $s$, and the number of patches in each dimension $k$. Then, the large map will be a square with size $B + s(k-1)$, and there will be $k^2$ patches, indexed by $(i, j)$, where $0 \le i, j < k$, such that patch $(i, j)$ spans height coordinates in $[is, is+B)$ and width coordinates in $[js, js+B)$. We denote patches of the satellite image as $X_{ij}$ and patches of the map conditioning as $Y_{ij}$. Then, we can note that due to the homogeneity of maps, the distribution $p(X_{ij}|Y_{ij})$ does not depend on $(i, j)$, so we can utilize the same pretrained conditional model on all the patches.

Our pretrained model for patches utilized the UNet architecture from  denoising\_diffusion\_pytorch (\href{https://github.com/lucidrains/denoising-diffusion-pytorch?tab=readme-ov-file}{url}). We pass the conditioning as input channels, so the UNet has 6 input channels (current image and conditioning), and 3 output channels. When training the model, we use the reparametrization of the network output from \citet{karras2022elucidating}. The UNet has 5 hidden layers with $[128, 256, 512, 1024, 1024]$ hidden units, respectively. We embed time with learned 32-dimensional sinusoidal embeddings. We use attention on the bottleneck layer, and on the upsampling and downsampling layers that are closest to it, with 8 attention heads and dimension 64 per head. In total, the base model has 270M parameters.

The coordinator model operates on the outputs of the base model UNets (i.e., we do not apply the reparametrization). We utilize the ViT architecture for the map coordinator. We split each patch into smaller patches used in ViT with size $2 \times 2$, and apply positional encodings that encode the position in the large image, and not the relative position in the patch. Since the $B \times B$-size patches may overlap, there could be ViT patches with the same positional encodings.

The ViT has depth 6, 6 attention heads, hidden size 384, and MLP ratio 4.0. It has 26M parameters. After all the transformer blocks, we unpatchify the tokens and extract 3 channels. Since we still have overlapping patches, we reconcile them by averaging the overlaps. Then, we apply the same reparametrization as the one that would have been applied to the outputs of the base model.

\subsubsection{Conditioning}

Here, we describe how ViT handles the conditional information (schematic map). We split a large $N \times N$ schematic map into $4 \times 4$ patches and pass them to ViT together with the patches of the model outputs. The ViT performs self-attention on all of the tokens; however, when performing MLP, we use different weights for tokens derived from model outputs and tokens derived from the conditioning.

\subsubsection{2D RoPE}

To improve the generalization capabilities of the ViT coordinator, we use rotary positional encodings from \citet{su2023roformer} (RoPE). RoPE are defined for the case when tokens have 1-dimensional positions. We do not add positional encodings explicitly at the beginning of the forward pass. Instead, in each layer of the transformer when we calculate keys or queries, we multiply them by a rotation matrix $R^d_{\Theta, m}$. Let $x_m$ be the $m$-th token, $d$ be its dimension, $\Theta$ be a parameter vector, $W_{\text{\{q, k\}}}$ be a learnable query/key matrix, respectively, and $f$ the function that computes the query/key of a token. Then, according to equation 14 from \citet{su2023roformer}:

\[
    f_{\text{\{q, k\}}}(x_m, m) = R_{\Theta, m}^dW_{\text{\{q, k\}}}x_m
\]

We generalize this method to the case when positions of tokens are two-dimensional. We do this by splitting our embeddings into two halves and applying the transform associated with the vertical coordinate on the first half, and the transform associated with the horizontal coordinate on the second half. Let $x_{nm}$ be a $2d$-dimensional token at position $(n, m)$. Then:

\[
    f_{\text{\{q, k\}}}(x_{nm}, n, m) =
    \begin{pmatrix}
        R_{\Theta, n}^d& 0 \\
        0& R_{\Theta, m}^d
    \end{pmatrix}W_{\text{\{q, k\}}}x_{nm}
\]

For 1D positional encodings, we have the property that the attention value between two tokens (i.e., the dot product between key and value) depends only on their coordinates and the relative position:

\[
    (f_k(x_n, n), f_q(x_m, m)) = g(x_n, x_m, n - m)
\]

By decomposing the vectors into two halves, it is easy to show that a similar statement holds for 2D RoPE:

\[
    (f_k(x_{ab}, a, b), f_q(x_{cd}, c, d)) = g(x_{ab}, x_{cd}, a - c, b - d)
\]

\subsubsection{Training procedure}

We train the coordinator for $100$ epochs, with a batch size of $16$. We sample the noise level $\sigma$ from a log-normal distribution with a mean of $-1$ and a standard deviation of $1.6$. We use the Adam optimizer with a learning rate of $3 \cdot 10^{-5}$.

\subsubsection{Sampler details}

We use the Heun sampler described in \ref{sec:app_exp_details} with $\sigma_{\max} = 20$, $\sigma_{\min} = 10^{-4}$, and $S_{\text{churn}} = 20.0$. For sigmas, we use the Karras schedule with $\rho = 7$ and $150$ timesteps.

\subsubsection{Evaluation details}

\begin{figure}[htbp]
    \centering
    \begin{minipage}{0.45\linewidth}
        \centering
        \includegraphics[width=\linewidth]{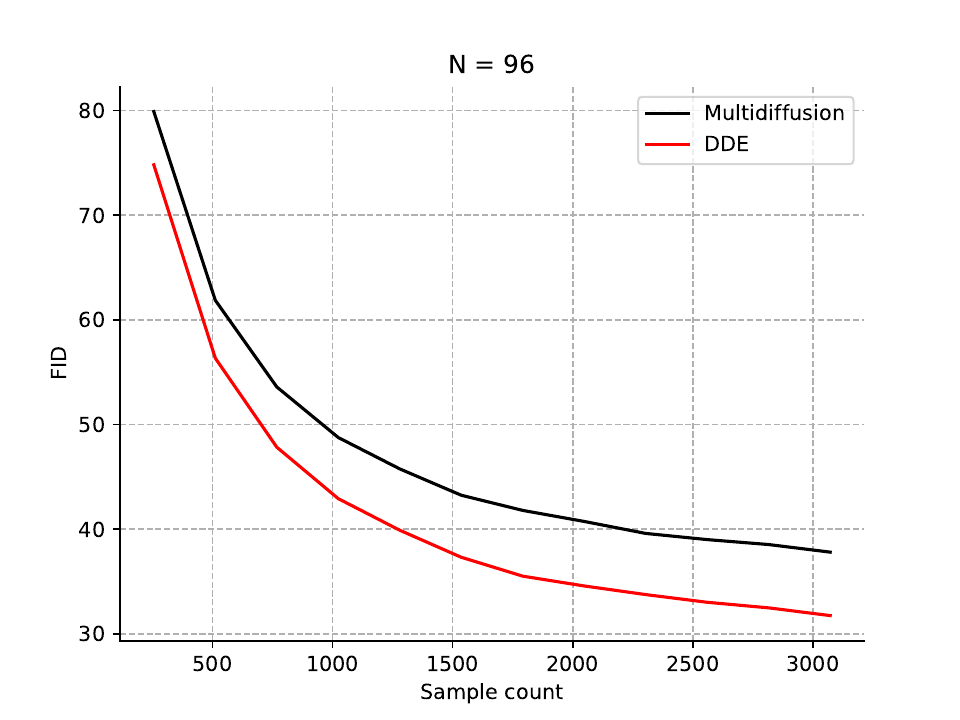}
    \end{minipage}
    \hfill
    \begin{minipage}{0.45\linewidth}
        \centering
        \includegraphics[width=\linewidth]{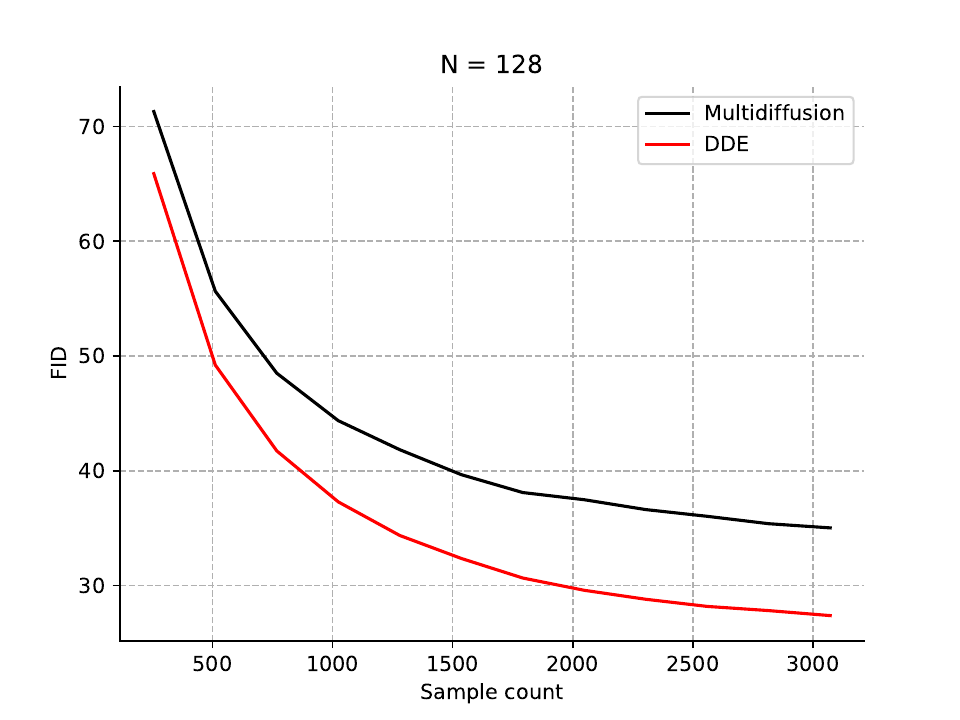}
    \end{minipage}
    \caption{Change of FID depending on the number of samples}
    \label{fig:fid}
\end{figure}

We use the CleanFID implementation of the FID metric \citep{parmar2021cleanfid}. When reporting the FID metric for a given model, we compare the whole satellite dataset, containing 7,300 images, with 3,072 generated samples. In Figure \ref{fig:fid}, we illustrate the dependency between the number of samples and the value of FID, showing that the comparison is consistent across the sample counts.

\newpage

\subsubsection{Samples}
\label{sssec:map_samples}

In each of the figures below, the columns indicate (from left to right) map conditioning, ground truth satellite image, and generated satellite image by DDE.

\begin{figure}[h!tbp]
    \centering
    \begin{minipage}{0.45\linewidth}
        \centering
        \includegraphics[width=\linewidth]{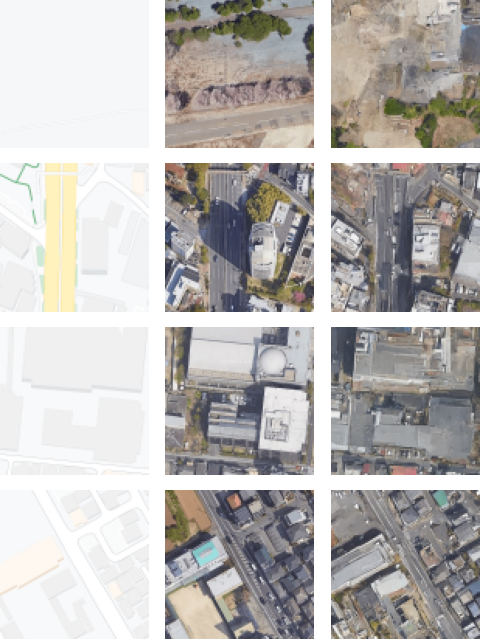}
        \caption{DDE samples of size $96 \times 96$}
    \end{minipage}
    \hfill
    \begin{minipage}{0.45\linewidth}
        \centering
        \includegraphics[width=\linewidth]{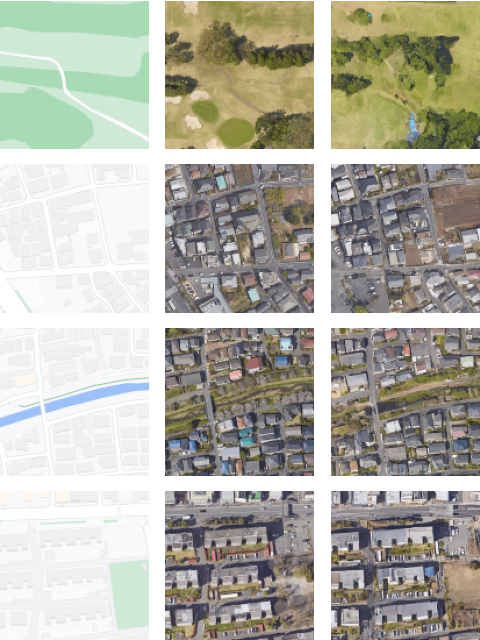}
        \caption{DDE samples of size $128 \times 128$}
    \end{minipage}
    \label{fig:maps96_128}
\end{figure}

\begin{figure}[h!]
    \centering
    \includegraphics[width=0.7\linewidth]{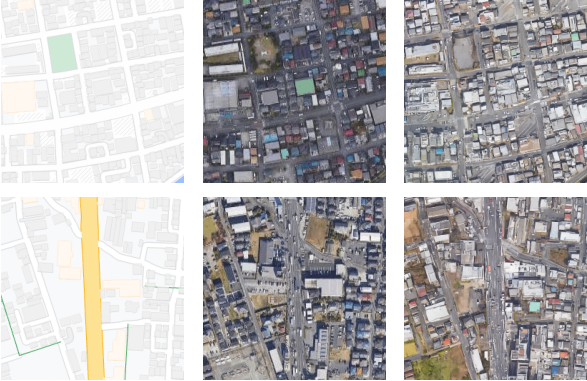}
    \caption{DDE samples of size $256 \times 256$}
    \label{fig:maps256}
\end{figure}

\end{document}